\documentclass[journal]{IEEEtran}

\usepackage{cite}
\usepackage[pdftex]{graphicx}
\usepackage{amsmath}
\usepackage{amssymb}
\usepackage{amsfonts}
\usepackage{algorithmic}
\usepackage{xcolor}
\usepackage{array}
\usepackage[caption=false,font=normalsize,labelfont=sf,textfont=sf]{subfig}
\usepackage{stfloats}
\usepackage{url}
\usepackage{verbatim}
\usepackage{color}
\usepackage{booktabs}
\usepackage{multirow}
\usepackage[ruled,linesnumbered]{algorithm2e} % options: linesnumbered, noend, etc.
\usepackage{soul}
\usepackage[pagewise]{lineno}

\usepackage{pdfpages} % Required to include PDF pages
\usepackage{pgffor}   % Required for the loop logic

\begin{document}

\newcommand{\T}[1]{\text{#1}}
\newcommand{\todo}[1]{\textcolor{blue}{TODO: #1}}
\newcommand{\ci}{\textcolor{blue}{[Citation Needed]}}

\title{Versatile Exoskeleton Control Supports Elderly Joint Energetics During Hip-Intensive Tasks}

\author{Jiefu~Zhang,~\IEEEmembership{Graduate Student Member, IEEE},
        Nikhil~V.~Divekar,~\IEEEmembership{Member, IEEE}, \\
        Chandramouli Krishnan,~\IEEEmembership{Member, IEEE},
        and~Robert~D.~Gregg,~\IEEEmembership{Senior~Member, IEEE}% <-this % stops a space
\thanks{This work was supported by the National Institute of Biomedical Imaging and Bioengineering of the NIH under Award Number R01EB031166. 
The content is solely the responsibility of the authors and does not necessarily represent the official views of the NIH.}
\thanks{The authors are with the Robotics Department, University of Michigan, Ann Arbor, MI 48109, USA. Contact: {\tt\footnotesize \{zjiefu, ndivekar, mouli, rdgregg\}@umich.edu}}}

\markboth{Journal of \LaTeX\ Class Files,~Vol.~14, No.~8, August~2015}%
{Shell \MakeLowercase{\textit{et al.}}: Bare Demo of IEEEtran.cls for IEEE Journals}



\maketitle

\begin{abstract}

Age-related mobility decline is frequently accompanied by a redistribution of joint kinetics, where older adults compensate for reduced ankle function by increasing demand on the hip. Paradoxically, this compensatory shift typically coincides with age-related reductions in maximal hip power. Although robotic exoskeletons can provide immediate energetic benefits, conventional control strategies have limited previous studies in this population to specific tasks such as steady-state walking, which do not fully reflect mobility demands in the home and community. Here, we implement a task-agnostic hip exoskeleton controller that is inherently sensitive to joint power and validate its efficacy in eight older adults. Across a battery of hip-intensive activities that included level walking, ramp ascent, stair climbing, and sit-to-stand transitions, the exoskeleton matched biological power profiles with high accuracy (mean cosine similarity 0.89). Assistance significantly reduced sagittal plane biological positive work by 24.7\% at the hip and by 9.3\% for the lower limb, while simultaneously augmenting peak total (biological + exoskeleton) hip power and reducing peak biological hip power and torque. These results suggest that hip exoskeletons can potentially enhance endurance through biological work reduction, and increase functional reserve through total power augmentation, serving as a promising biomechanical intervention to support elderly mobility.

\end{abstract}

\begin{IEEEkeywords}
exoskeletons, biomechanics, control
\end{IEEEkeywords}

% \textbf{Keywords: }exoskeletons, aging, control, biomechanics

% \fi
% \IEEEpeerreviewmaketitle

\section{Introduction}
\IEEEPARstart{M}{obility} impairments are becoming increasingly prevalent as the population ages, threatening independence and diminishing quality of life \cite{miller2001influence, metz2000mobility}. A primary driver of this decline is the loss of lower-limb muscle power, which drops by up to 50\% in older adults \cite{harbo2012maximal, alcazar2023ten, runge2004muscle} and is strongly correlated with their functional status \cite{foldvari2000association}. The hip is particularly vulnerable, exhibiting maximal power reductions of 31--39\% compared to young adults \cite{dean2004age}. Paradoxically, older adults compensate for reduced distal ankle function by shifting mechanical demand proximally to the hip \cite{delabastita2021distal, devita2000age}. This ``distal-to-proximal shift'' forces the weakened hip to contribute a greater proportion of mechanical work during activities of daily living (ADLs) \cite{farris2012mechanics}. The resulting ``demand-capacity mismatch'' depletes the joint's functional reserve, rendering older adults prone to fatigue \cite{arnett2008aerobic, enoka2008muscle} and constraining performance in high-demand tasks such as stair climbing \cite{graf2005effect, wilken2011role, bassey1992leg}. Therefore, interventions that offload this compounding energetic burden are critical to preserving physiological reserve and mitigating mobility loss.

Current interventions fall into device-based and non-device-based categories, both of which have limited impact on the demand-capacity mismatch in older adults. Non-device interventions, such as physical training aimed at improving muscle strength/power and thus functional capacity, have shown clear benefits for improving hip function \cite{portegijs2008effects, kim2013unsupervised, beijersbergen2017hip}, but they typically lack immediate benefits \cite{hunter2004effects} and place substantial demands on adherence and training intensity that may be unattainable for individuals with significant comorbidities (e.g., pain, limited range of motion, or unstable cardiovascular conditions), contributing to high dropout rates \cite{fisher2017minimal}. Moreover, these interventions generally only delay or partially reverse muscular deterioration, without fully restoring power deficits resulting from natural aging \cite{booth2011lifetime, bickel2011exercise}. Passive assistive devices such as crutches or walkers can enhance balance and reduce lower-limb joint load \cite{joyce1991canes}, but their inability to inject energy makes them biomechanically ineffective during power-demanding tasks \cite{thys1996energy}. Furthermore, these mechanical aids are impractical on complex terrain such as stairs.

Powered hip exoskeletons can augment hip mechanical power and offload the joint, offering a promising way to address the age-related demand-capacity mismatch.
While devices like the Samsung GEMS hip exoskeleton have been used for rehabilitation in older adults \cite{jayaraman2022modular}, this application shares the key limitations of the exercise-based strategies described above. Instead, enabling immediate performance gains, such as improved speed and metabolic efficiency during level walking \cite{leeGaitPerformanceFoot2017, leeWearableHipAssist2017} and stair climbing \cite{kimWearableHipassistRobot2018}, presents a more compelling case for real-world mobility. A research prototype soft hip exoskeleton also demonstrated metabolic benefits in both young and older adults during walking \cite{tricomiSoftRoboticShorts2024}. However, the exoskeleton controllers in these studies were not versatile across a variety of activities and did not directly address the joint demand-capacity mismatch. In \cite{leeGaitPerformanceFoot2017, leeWearableHipAssist2017, tricomiSoftRoboticShorts2024}, assistive torque was determined by first estimating the gait phase and then applying a predetermined torque pattern, which is ill-suited for varying, unstructured ADLs. The controller in \cite{kimWearableHipassistRobot2018} directly maps kinematics to assistive torque but was designed solely for stair climbing. 
While these results are promising, real-world use requires a controller that predictably adds power when needed while attenuating assistance when unnecessary or potentially destabilizing. For example, stair descent typically requires little hip assistance but is strongly associated with falls in older adults \cite{nevitt1991risk}; mistimed or excessive torque can perturb balance and increase fall risk.

To effectively address age-related power deficits in daily life, exoskeletons must provide meaningful energetic support while adapting and interacting predictably. Although versatile (i.e., \emph{task-agnostic}) controllers have been developed to produce biomimetic assistance across multiple tasks \cite{van2025ai}, they have not yet demonstrated joint-level energetic assistance in older adults. Deep learning-based approaches trained on healthy-population datasets (e.g., \cite{molinaro2024a}) have shown versatile assistance and energetic benefits in young, healthy participants, but these black-box approaches lack interpretability and predictability in the presence of untrained gait pathologies.

Model-based approaches such as \cite{lv2018design} shape the closed-loop human--exoskeleton system dynamics to synthesize assistive torque. This framework provides an analytical structure that constrains energy injection for theoretical stability guarantees. By using a large set of nonlinear basis functions, later works \cite{zhang2023optimal, lin2024modular, walters2025optimal} improved the matching of the assistive torques with biological torque profiles for hip, knee, and ankle exoskeletons. While these basis functions were mathematically well-founded, they were biomechanically unintuitive, limiting customizability based on subjective feedback. Consequently, an alternative task-agnostic control philosophy has emerged using energetically bounded, physically meaningful components (e.g., nonlinear springs, dampers, and gravity/inertial compensation) modulated by phase- and task-sensitive signals \cite{divekar2024a}. Initially designed for knee exoskeletons, this approach successfully mitigated fatigue in unimpaired users \cite{divekar2024a} and assisted motor control deficits associated with neurological impairments \cite{divekar2025customizing} across multiple tasks.

\begin{figure}[tbp]
  \centering
    \includegraphics[width=0.85\linewidth]{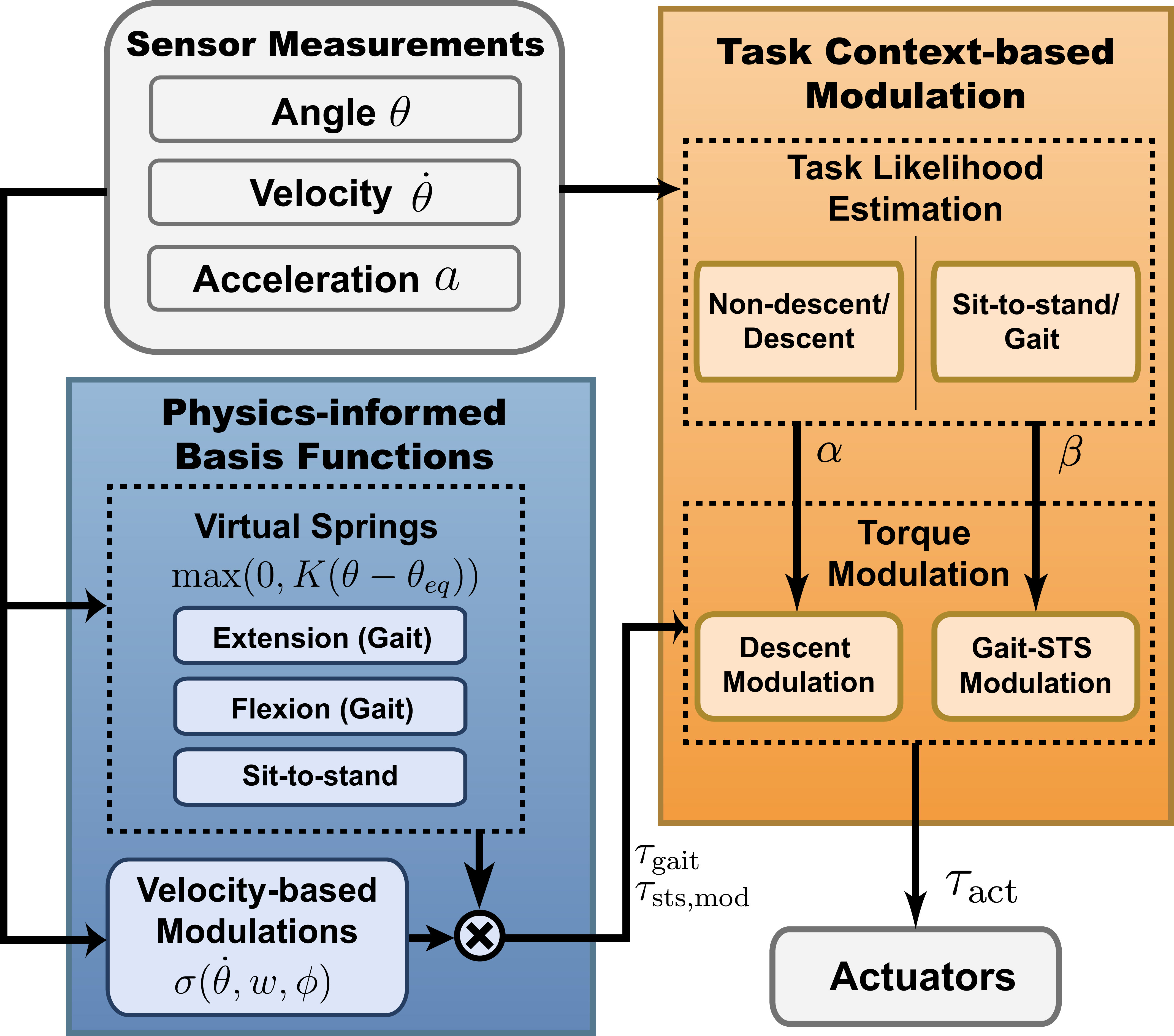}
  \caption{Overview of the physics-informed task-agnostic exoskeleton controller structure. Sensor measurements are used to generate velocity-modulated virtual spring torque bases, which are subsequently modulated based on task context to generate the output torque.}
  \label{fig:schematic}
  \vspace{-4mm}
\end{figure}

Following this control philosophy, this paper develops a task-agnostic bilateral hip exoskeleton controller using velocity-modulated virtual springs to support hip joint energetics in older adults. This approach is motivated by prior evidence for the energetic benefits of spring-based control \cite{shafer2023} and the fact that velocity modulation naturally couples assistive torque to joint velocity and thus assistive power \cite{dingEffectTimingHip2016}. While we previously demonstrated the feasibility of velocity-modulated virtual springs for reducing biological hip torque and associated osteoarthritic pain \cite{zhang2025task}, this pilot study did not address joint energetics for elderly users. This preliminary controller also lacked a mechanism for ensuring transparent interaction during descent tasks, which is a known user preference \cite{lin2024modular}. Here we reformulate the controller into a hierarchical architecture shown in Fig.~\ref{fig:schematic}: a physics-informed basis layer generates assistive torques based on kinematic inputs, and a modulation layer scales these torques by task context, including a torque attenuation mechanism for low-demand, high-risk descent tasks.

To test our hypothesis that task-agnostic hip assistance would induce holistic joint energetic benefits in older adults, we implemented our controller on a bilateral hip exoskeleton \cite{nesler2022} and evaluated it in eight participants across diverse ADLs with and without assistance. We focus on joint-level energetics rather than metabolic cost, as they are more strongly linked to functional status in older adults \cite{foldvari2000association}, and better reflect performance in short-bout, high-demand activities such as stair ascent and sit-to-stand \cite{wilken2011role, bassey1992leg}. Accordingly, our primary outcome measure was biological hip positive work during hip-intensive activities, and our secondary outcome measures were biological lower-limb positive work (hip + knee + ankle), peak total hip power (biological + exoskeleton), and peak biological hip power, across hip-intensive activities. Overall, this work provides (i) a task-agnostic control paradigm targeting age-related joint energetic deficits, and (ii) a comprehensive joint-level biomechanical evaluation in older adults across diverse ADLs, extending beyond prior studies in this population that examined only one or two tasks. In conclusion, this work supports task-agnostic powered hip exoskeletons as a potential biomechanical intervention for age-related mobility deficits.

\section{Controller Design and Hardware Implementation}

\begin{figure}[htbp]
  \centering
  \begin{tabular}{ll}
    \includegraphics[width=0.40\linewidth]{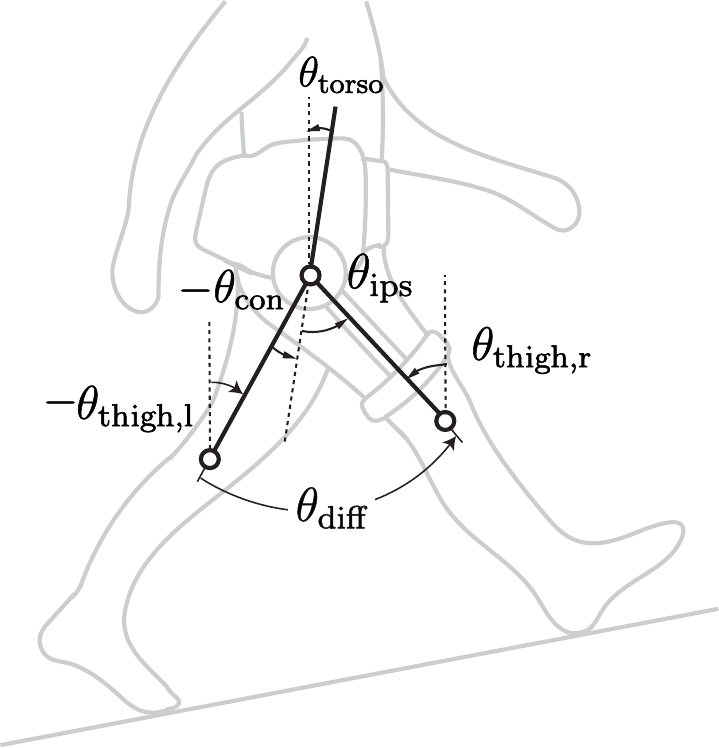}
    &
    \includegraphics[width=0.31\linewidth]{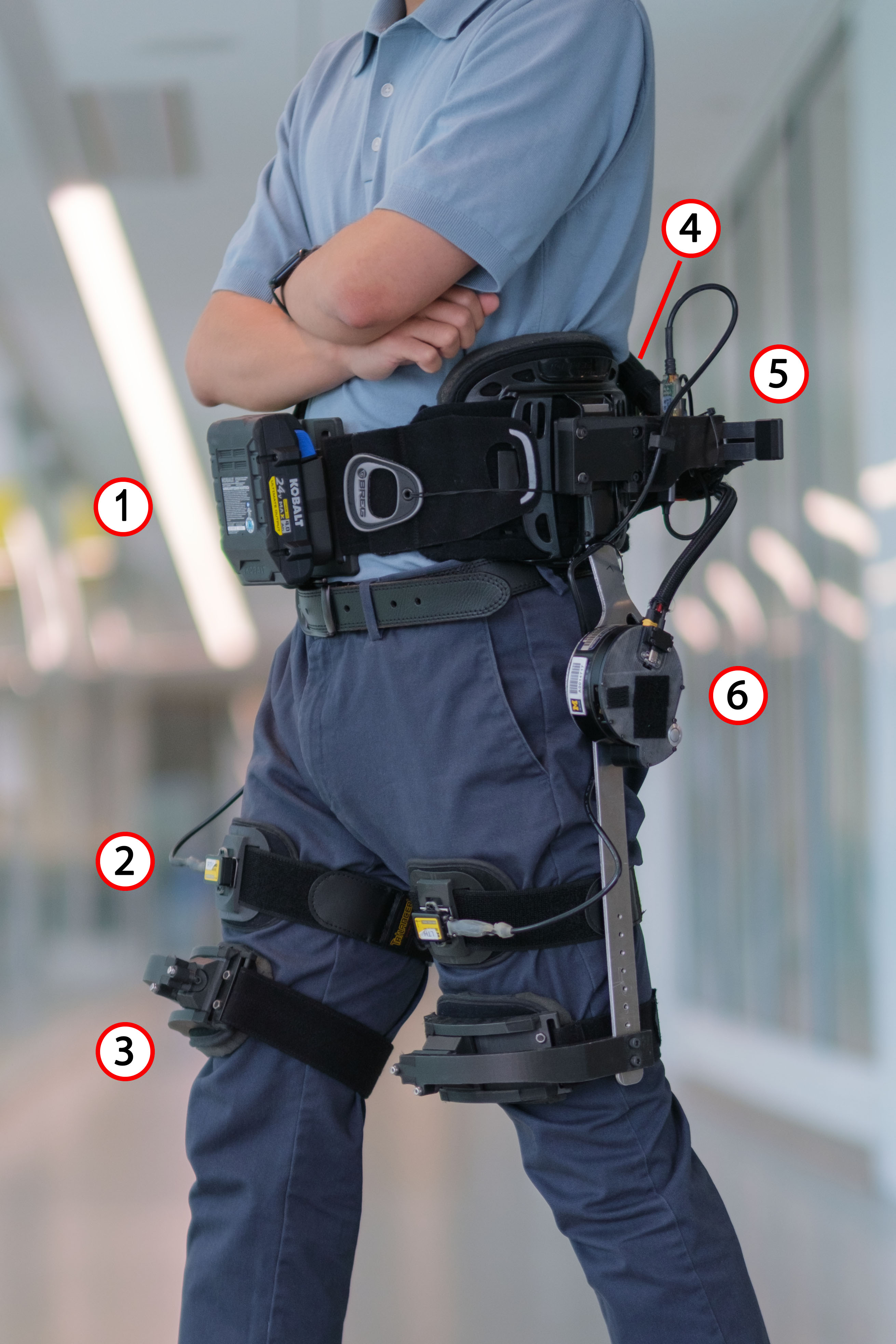}
  \end{tabular}
  \caption{\textbf{Left: } Definitions of angles used in the controller. Black-outlined circles represent hip and knee joint centers. $\theta_{\text{torso}}, \theta_{\text{thigh,l}}, \theta_\T{thigh,r}, \theta_{\text{ips}}$, and $\theta_{\text{con}}$ denote the angles of the torso, left and right thigh, ipsilateral hip, and contralateral hip, respectively. $\theta_\T{diff}$ represents the inter-thigh angle difference.
  Hip flexion is defined as the positive direction. Figure reproduced from \cite{zhang2025task} with permission. \textbf{Right: } Picture of the modified M-BLUE hip module: (1) Battery, (2) Inertial measurement unit (IMU), (3) Thigh interface, (4) Electronics Box, (5) Post-op Brace, (6) Actuator. For clarity only one of the IMUs, actuators, and thigh interfaces are labeled.}
  \label{Fig:model}
\end{figure}

This section presents our biomechanics-based, task-agnostic controller for a bilateral hip exoskeleton (Fig.~\ref{fig:schematic}). We first describe the physics-informed basis functions and velocity-based modulations that generate interpretable assistance torques, corresponding to the ``Physics-informed Basis Functions'' layer in Fig.~\ref{fig:schematic}. We then introduce a task context-based modulation layer that scales this torque according to the estimated likelihood of different task types, enabling task-adaptive assistance (the ``Task Context-based Modulation'' layer in Fig.~\ref{fig:schematic}). Next, we detail our parameter design procedure across tasks and participants. Finally, we describe the hardware implementation of the controller on a bilateral hip exoskeleton.

\subsection{Physics-Informed Bases and Velocity-Based Modulation}
\label{sec:controller design}

The physics-informed basis layer generates assistive torques based on kinematic inputs from sensor measurements. To support older adults' joint energetics while remaining robust to varying kinematics, we choose three velocity-modulated virtual torsion springs as the physics-informed basis functions of our controller: two gait springs that generate $\tau_{\T{ext}}$ and $\tau_{\T{flex}}$---extension and flexion torques for gait tasks (e.g., walking and stair climbing), and one sit-to-stand (STS) spring that generates $\tau_{\T{sts}}$---extension torque for STS transitions. Fig.~\ref{Fig:model} (left) provides the definition of angles used in this controller.

The extension and flexion torques from the respective gait springs are determined as
\begin{equation}
  \begin{split}
      \tau_{\T{ext}} &= \min(0, k_{\T{ext}} \cdot (\theta_{\T{ips}} - \theta_{\T{ext,eq}}))\  \\
      \tau_{\T{flex}} &= \max(0, k_{\T{flex}} \cdot (\theta_{\T{flex,eq}} - \theta_{\T{ips}}))\ 
  \end{split}
\end{equation}
where $\theta_{\T{ips}}$ is the ipsilateral hip angle, $k_{\T{ext}}, k_{\T{flex}}$ are the constant spring stiffness coefficients of the extension and flexion springs, and $\theta_{\T{ext,eq}}, \theta_{\T{flex,eq}}$ are the equilibrium angles of the extension and flexion springs, respectively. The $\min(\cdot)$ and $\max(\cdot)$ functions ensure that the springs generate torques only in their intended directions (i.e., positive for flexion).

Due to differences between gait tasks and STS transitions, the STS spring is designed separately from the gait springs. The extension torque from the STS spring is given by
\begin{equation}
  \tau_{\T{sts}} = \min (0, -k_{\T{sts}} \cdot \theta_{\T{thigh}})\ 
\end{equation}
where $\theta_{\T{thigh}}$ is the ipsilateral thigh angle, as defined in Fig.~\ref{Fig:model}. To generate extension restoring torques only when the thigh is flexed, we make the spring unidirectional and set the equilibrium angle at $\theta_{\T{thigh}} = 0^{\circ}$.

Although virtual springs capture key features of hip biomechanics \cite{shafer2023}, they generate torques solely from joint angle. Since similar hip angles can occur in different gait phases and activities with different biological moment demands, angle-only mappings cannot disambiguate context to tailor the torque profile. This limitation is especially consequential for our energetic objective, which requires precise power timing---maximizing positive exoskeleton power by aligning peak assistive torque with periods of high angular velocity \cite{dingEffectTimingHip2016}. To overcome this limitation while preserving biomechanical interpretability, we modulate the virtual-spring torque using joint velocity, which provides phase sensitivity and enables independent shaping of torque magnitude beyond angle alone. This also allows the controller to supplement the two positive hip power peaks: the H1-S peak during early-to-mid stance, and the H3-S peak during late-stance to early-swing \cite{winter2009biomechanics}. 

A sigmoid function $\sigma$ is defined as
\begin{equation}
  \label{eq:sigmoid}
  \sigma(x; w, \phi) = \frac{1}{1 + e^{-w x + \phi}} .
\end{equation}
where $x$ is the input, and $w$ and $\phi$ are constant parameters that control the slope and horizontal shift of the sigmoid. For $x \in \mathbb{R}$, $\sigma$ smoothly maps to $[0,1]$, providing desirable properties for velocity modulation. With angular velocity as the input, tuning $w$ and $\phi$ adjusts the sensitivity to velocity and thus shapes the amplitude of the assistive torque profile.

The velocity-based modulation factors for the gait springs can be determined by
\begin{equation}
  \begin{split}
    \eta_{\T{ext}} &= \sigma (\dot{\theta}_{\T{ips}}, w_{\T{ext}}, \phi_{\T{ext}})\  \\
    \eta_{\T{flex}} &= \sigma (\dot{\theta}_{\T{ips}}, w_{\T{flex}}, \phi_{\T{flex}})\ 
  \end{split}
\end{equation}
where $\dot{\theta}_{\T{ipsi}}$ is the ipsilateral hip angular velocity, and $w_{\T{ext}}, w_{\T{flex}}, \phi_{\T{ext}}, \phi_{\T{flex}}$ are parameters to be determined. 
The total torque from the gait springs after velocity-based modulation can then be expressed as
\begin{equation}
\label{eq:vel_mod}
  \tau_\T{gait} = \eta_\T{ext} \cdot \tau_\T{ext} + \eta_\T{flex} \cdot \tau_\T{flex}\ .
\end{equation}

Similar to the gait springs, the STS spring uses hip angular velocity to shape phase-dependent assistance. The velocity-based modulation factor is determined as
\begin{equation}
    \eta_{\T{sts,vel}} = \sigma(\dot{\theta}_{\T{ips}}, w_{\T{vel}}, \phi_{\T{vel}})\ .
\end{equation}
By tuning the slope and offset, $\eta_{\T{sts,vel}}$ promotes assistance during rising (hip extension) and reduces assistance during sitting to avoid uncomfortable bracing. 

For timely support during the rapid extension phase of rising, the velocity-based modulation of the STS spring must be highly sensitive to hip angular velocity. However, this can also cause unintended responses to small, incidental hip motions while the user remains seated (e.g., posture shifts), producing undesired torque. To preserve fast assistance onset during standing while suppressing extension torque during seated postures, we introduce an additional state-awareness term. Specifically, we further scale the STS spring by a sigmoid function of torso angle, as initiating STS typically involves a substantial anterior trunk lean (and remaining seated does not). This yields the torso-angle-based modulation factor
\begin{equation}
    \eta_{\T{sts,torso}} = \sigma(\max(0, \theta_{\T{torso}}), w_{\T{torso}}, \phi_{\T{torso}})\ .
\end{equation}
The $\max(0,\theta_{\T{torso}})$ term ensures the STS spring engages only when $\theta_{\T{torso}}>0$ (i.e., when the torso leans forward), suppressing velocity-triggered extension torque due to seated movements. Together, the modulated torque for STS transitions is
\begin{equation}
    \tau_\T{sts,mod} = \tau_\T{sts} \cdot \eta_\T{sts,vel} \cdot \eta_\T{sts,torso}\ .
\end{equation}

\subsection{Task Context-Based Modulation}
\label{sec:gait length mod}

Velocity-modulated virtual springs provide intuitive, velocity-sensitive torque bases. Yet, the basis layer alone cannot guarantee both strong assistance during hip-demanding activities and transparent (i.e., low torque) interactions during low-demand yet high-risk tasks (e.g., stair descent). We therefore add a task-context modulation layer that scales the basis-layer torques using kinematic context, enabling seamless assistance across activities while tapering excessive torques when mechanical transparency is prioritized.

\subsubsection{Descent Modulation}
\label{sec:des mod}

Compared to activities such as level walking or ramp ascent that require substantial positive hip power, descent tasks (e.g., ramp or stair descent) typically demand relatively low hip effort \cite{farris2012mechanics} and thus less exoskeleton assistance, yet they are also associated with elevated fall risk in older adults \cite{nevitt1991risk}. During descent, the early-to-mid stance period predominantly requires net negative work, and thus lacks the characteristic H1-S peak in hip-intensive tasks which the physics-informed layer is designed to support. As a result, the controller can generate excessive hip extension torque during descent, which could be uncomfortable and may cause potentially risky, premature foot contact during late-swing. To alleviate this, we propose an improved descent attenuation mechanism that naturally and consistently reduces extension torque during descent, improving torque predictability and supporting more transparent interactions in older adults. This attenuation mechanism prevents undesirable positive power generation at the hip, which differs from prior work that improved transparency during descent by tapering negative power-generating bracing torques \cite{lin2024modular}.

Inter-thigh angle difference $\theta_\T{diff}$ at HS provides informative task-context cues, which have been used for real-time gait classification in exoskeleton and prosthesis applications \cite{jang2015online, posh2025task}. Ascent activities typically exhibit a larger inter-thigh angle difference at HS, reflecting increased hip flexion for ground clearance, whereas descent activities often show a smaller difference because step length is shortened to accommodate higher braking demands \cite{sun1996influence}. Building on these observations, we use inter-thigh angle difference at HS to contextually modulate hip extension assistance: maintaining strong power delivery during high-demand tasks while attenuating excessive extension torque during high-risk descent tasks.

Since this strategy relies on HS-specific kinematics, accurate and robust real-time HS detection is required, especially during descent. Foot sensors offer robust foot contact detection \cite{pappas2001reliable}, but are less suitable for hip exoskeletons. Data-driven IMU-based methods have demonstrated strong performance \cite{wei2023gait,garcia2022adaptive}, but can be sensitive to sensor placement and orientation \cite{anwary2018optimal}, which are often constrained on exoskeletons. To streamline implementation and ensure compatibility with the M-BLUE hip exoskeleton \cite{nesler2022}, we use a rule-based detector based solely on the existing thigh and pelvis IMUs. During walking, HS is identified by a characteristic spike in thigh-normal linear acceleration measured by the thigh IMU. During stair descent, reduced thigh flexion at HS weakens this signature, but the pelvis IMU, in contrast, captures the larger center-of-mass deceleration. Combining these complementary signals yields a simple HS detector that generalizes across locomotion tasks while balancing accuracy and practicality. The full algorithm is detailed in the Supplemental Material.

Once a HS is detected, we compute the descent modulation factor $\alpha$ from the inter-thigh angle difference at HS. As a safeguard, we first check whether the left and right thigh angles at HS, $\theta_\T{thigh,l}$ and $\theta_\T{thigh,r}$, fall outside a reasonable range $[\theta_\T{thigh,m}, \theta_\T{thigh,M}]$. If deviating from this range, we set $\alpha = 0$ to disable descent attenuation. This prevents unintended attenuation when HS occurs with large thigh flexion or extension---characteristics of non-descent activities \cite{posh2025task}. Otherwise, $\alpha$ is computed from the HS inter-thigh angle difference as
\begin{equation}
\label{eq:alpha}
\alpha = \sigma\left(|\theta_\T{diff}|, w_\T{step}, \phi_\T{step}\right)\ 
\end{equation}
where $w_\T{step}$ and $\phi_\T{step}$ determine the sensitivity of the modulation to thigh angle differences at HS. From (\ref{eq:alpha}), a smaller inter-thigh angle difference at HS yields a larger $\alpha$. 

Then $\alpha$ is used to modulate the resulting torque $\tau_\T{gait}$ from the gait springs in (\ref{eq:vel_mod}) as follows:
\begin{equation}
  \label{eq:tau_gait}
  \tau_\T{gait,mod} = (1-\lambda\alpha) \cdot \min(0, \tau_\T{gait}) + \max(0, \tau_\T{gait})\ ,
\end{equation}
where $\lambda$ is a fixed parameter that adjusts the modulation strength. As shown in (\ref{eq:tau_gait}), only extension torque ($\tau_\T{gait}<0$) is attenuated, while flexion torque remains unchanged. Thus, during descent, the modulation selectively suppresses potentially risky excessive extension torque (that supports H1-S for hip-intensive activities), while preserving beneficial flexion assistance (that supports H3-S). Overall, this new design provides task-adaptive scaling of assistance while maintaining consistent controller behavior across activities, improving predictability and user confidence.

We also include a reset rule that clears descent attenuation when the user returns to quiet standing. After detecting a sustained standing posture for $T_\T{wait}$ (using the standing indicator $\beta$, introduced later), we linearly ramp the descent modulation factor $\alpha$ back to $0$ over $T_\T{decay}$, restoring nominal assistance for the subsequent movement (e.g., stand-to-walk).

\subsubsection{Gait-STS Modulation}
\label{sec:sts int}

Now we consider gait-STS modulation, which blends $\tau_\T{gait,mod}$ and $\tau_\T{sts,mod}$ based on bilateral kinematic symmetry. Similar to \cite{zhang2025task}, we define the modulation factor $\beta$ and use it to scale the contributions from the gait and STS springs, since bilateral symmetry is a strong differentiator of walking and STS activities. Specifically, $\beta$ is calculated as
\begin{equation}
\label{eq:sym}
    \beta = \sigma\left(|\theta_\T{diff}|, w_\T{sc}, \phi_\T{sc} \right) \cdot \T{step}\left(\dot{\theta}_\T{thre} - |\dot{\theta}_\T{diff}| \right )\ 
\end{equation}
where $\dot{\theta}_\T{diff}$ is the inter-thigh velocity difference,  $\dot{\theta}_\T{thre}$ is the bilateral velocity-difference threshold, and $w_\T{sc}$ and $\phi_\T{sc}$ are parameters to be determined. When $w_\T{sc}$ and $\phi_\T{sc}$ are negative, the sigmoid term in (\ref{eq:sym}) increases as bilateral angle difference decreases. This allows the controller to exploit bilateral hip-angle symmetry, where bilateral joint angles are nearly identical, to identify standing or sitting postures. The step function further enforces that $\beta$ is nonzero only when the bilateral velocity difference falls below $\dot{\theta}_\T{thre}$, preventing false positives when hip angles are symmetric but the motion is not (e.g., during walking). In addition, we set $\beta=1$ when both $\theta_\T{thigh,l}$ and $\theta_\T{thigh,r}$ exceed a preset flexion threshold, indicating a seated posture; this suppresses gait-module torque when a user is seated and moving their legs, which could otherwise violate the symmetry assumptions and spuriously increase $\beta$. To ensure smooth transitions, we filter $\beta$ using an exponential moving average filter \cite{klinker2011exponential} with smoothing coefficient $0.1$ \cite{jang2015online}. Finally, since $\beta$ depends only on symmetry, it also serves as a reliable indicator of quiet standing; we leverage this property to trigger the nominal-behavior reset described above, improving controller robustness.

Subsequently, torques from the gait and STS modules are combined as a convex combination:
\begin{equation}
  \tau_\T{act} = \beta \cdot \tau_\T{sts,mod} + (1-\beta) \cdot \tau_\T{gait,mod}\ .
 \end{equation}

\subsection{Parameter Design}

To achieve task-agnostic, biomimetic assistance while improving energetic benefits and comfort, we adopt a two-phase parameter design process inspired by \cite{divekar2024a}. In Phase~1, we formulate parameter design as an optimization problem using normative lower-limb kinematics and hip joint moments from open-source multi-activity datasets \cite{camargoComprehensiveOpensourceDataset2021, laschowski2021simulation}, specifically level walking at 0.85 m/s and 1.15 m/s (LG 0.85/1.15), ramp ascent/descent at 5.2$^\circ$ and 11$^\circ$ inclinations (RA/RD 5.2/11), and ascending/descending stairs with 5 and 7 in step height (SA/SD 5/7). Given normative input kinematics across these ADLs, we optimize the controller's parameters to  minimize the discrepancy between estimated assistive torque and normative biological hip moment, which also minimizes joint power discrepancies as joint velocities are fixed by the input kinematics. The parameters that were optimized include the slopes $w_\T{ext}, w_\T{flex}$ and offsets $\phi_\T{ext}, \phi_\T{flex}$ in the velocity-based modulation, and the equilibrium angles $\theta_\T{ext,eq}, \theta_\T{flex,eq}$ of the gait springs. We use the same objective function as \cite{zhang2025task}, comprising: i) torque-tracking error with activity-dependent weights to emphasize hip-demanding tasks, ii) a penalty on nonzero static torque (evaluated at zero kinematics) to avoid discomfort during quiet standing, and iii) a sign-mismatch penalty to discourage assistance in the wrong direction. We solve the optimization in MATLAB R2023b using the ``\texttt{fmincon}'' function with physically reasonable lower and upper bounds. 

To assess optimization results, we define the cosine similarity (SIM) as a measure of matching between estimated torque $\tau_\T{est}$ and biological moment $\tau_\T{bio}$:
\begin{equation}
    \text{SIM} = \frac{\tau_\T{est} \cdot \tau_\T{bio}}{\| \tau_\T{est}\|_2 \cdot \|\tau_\T{bio} \|_2},
\end{equation}
where $\|\cdot \|_2$ represents the vector 2-norm. The SIM between $\tau_\T{est}$ and $\tau_\T{bio}$ for the activities included in the in silico optimization is reported in Table~\ref{tab:silico_SIM}, where 1 indicates perfect similarity.

While this optimization yields good torque matching in simulation, normative datasets do not capture human-exoskeleton interaction effects that may be critical for real-world comfort and benefit. Because our biomechanics-informed controller has interpretable parameters that can be adjusted to target specific interaction goals \cite{divekar2025customizing}, we performed a second, user-in-the-loop refinement step. In Phase~2, we used the Phase~1 solution as a \textit{warm start} and fine-tuned the parameters based on subjective feedback from one young healthy user (male, 37~years old, 79~kg, 1.78~m) during level and inclined treadmill walking, with the goal of achieving natural, helpful interactions. All tuning was completed prior to the main study; parameters were held fixed for the subsequent experiments with older adults.

\begin{table*}[htbp]
\centering
\footnotesize
\setlength{\tabcolsep}{4pt}
\renewcommand{\arraystretch}{1.2}
\caption{Cosine similarity (SIM) between exoskeleton power and normative biological hip power (SIM = 1 indicates perfect similarity).}
\label{tab:silico_SIM}
\begin{tabular}{c c c c c c c c c c c}
\toprule
\textbf{Activity}
& \textbf{LG 0.85}
& \textbf{LG 1.15}
& \textbf{RA 5.2}
& \textbf{RA 11}
& \textbf{SA 5}
& \textbf{SA 7}
& \textbf{RD 5.2}
& \textbf{RD 11}
& \textbf{SD 5}
& \textbf{SD 7} \\
\midrule
\textbf{SIM}
& 0.9751
& 0.9805
& 0.9815
& 0.9802
& 0.9661
& 0.9672
& 0.7029
& 0.7785
& 0.7567
& 0.6843 \\
\bottomrule
\end{tabular}
\vspace{-3mm}
\end{table*}

\subsection{Hardware Implementation}

The controller was implemented on the bilateral hip exoskeleton in Fig. \ref{Fig:model} (right), an improved hip module of the M-BLUE system \cite{nesler2022} with details in Supplemental Material. The exoskeleton combines commercial hip braces (Breg T-Scope Hip) with two highly backdrivable actuators (T-motor AK80-9 with Dephy FASTER driver), each providing up to 25~Nm peak torque with low reflected inertia (92.1~$\text{kg}\cdot\text{cm}^2$). For safety and thermal considerations, actuator torque was limited to 22~Nm during experiments. Three IMUs (Microstrain 3DM-GX5 AHRS) were attached to the thigh bands (left/right) and waist belt (torso) to measure sagittal thigh, torso, and hip angles and angular velocities in real time. The control algorithm ran on a Raspberry Pi 5 (8 GB RAM) at 250 Hz. Hip angular velocity and commanded torque were filtered with second-order Butterworth low-pass filters (10 Hz and 5 Hz cutoff, respectively). The system was powered by a 24~V, 4~Ah lithium-ion battery (Kobalt 24V), for a total mass of 4.6~kg.

\section{Experimental Methods}

To study our hypotheses, we designed a study consisting of a multi-activity test, approved by the Institutional Review Board at the University of Michigan (HUM00201957). The study was registered as clinical study at clinicaltrials.gov (NCT05240014).
To determine the required sample size, we performed an a priori power analysis using data from pilot participants, with biological hip positive work reduction as the primary outcome. To achieve 90\% statistical power (see Supplemental Material), we enrolled eight elderly participants (4 males, 4 females, age: 72 $\pm$ 4.9~years old, body weight: 69.1 $\pm$ 7.0~kg, height: 1.72 $\pm$ 0.04~m). All participants gave written informed consent prior to participation and were required to wear a safety harness throughout the study to prevent falls.

\subsection{Experimental Design}

The experiment comprised an acclimation session and a data collection session on separate days. During the acclimation session, the exoskeleton was fitted, and participants practiced performing the experiment activities with the assistance. The level of assistance (LoA) started low and ramped up until the target LoA was reached (the maximum extension torque across activities reaching 22 Nm). We also provided instructions to help participants benefit the most from the assistance. 

During the data collection session, participants performed a variety of activities. Participants walked on a Bertec instrumented treadmill with a level slope at 1.0 (LG 1.0) and 1.2~m/s (LG 1.2), a 5$^{\circ}$-inclined (RA 5) and 5$^{\circ}$-declined (RD 5) slope at 1.0~m/s, and a 10$^{\circ}$-inclined (RA 10) and 10$^{\circ}$-declined (RD 10) slope at 0.8~m/s. Participants also performed stair ascending (SA) and descending (SD) on a 5-step staircase with 7~in (178~mm) step height, and STS transitions from a knee-height chair. All activities were performed both with the powered exoskeleton (EXO) and without the exoskeleton (No EXO), and the sequence of conditions was randomized. Motion capture data were collected at 250 Hz using a 27-camera motion capture system (Vicon Motion Systems), and ground reaction force (GRF) data were collected at 1000~Hz using force plates embedded in the treadmill for LG, RA, RD (Bertec), stairsteps for SA, SD (Kistler), and ground for STS (AMTI Inc.). The motion capture data and force data were synchronized using a Vicon Lock Sync Box. During EXO condition trials, we used a Delsys Trigger Module to initiate motion capture and force recording, and simultaneously sent a trigger to a custom Raspberry Pi, which relayed a synchronization signal over Wi-Fi to the exoskeleton Raspberry Pi for offline alignment of motion capture, force, and exoskeleton log data.

\subsection{Data Processing and Analysis}

\subsubsection{Biological Analysis}

All biomechanical analyses were performed in OpenSim \cite{delp2007opensim, seth2018opensim} using the \textit{gait2392} model. For each participant, a standing calibration trial was used to scale the generic model and place virtual markers to match experimental marker locations. Pelvis tracking under the exoskeleton was achieved with a custom inverted T-shaped marker cluster mounted on a sacrum plate (with pass-through slots for the waist belt/clothing). 
During the EXO condition, we omitted greater trochanter markers due to occlusion. The same scaled and marker-placed model was used for both conditions, with additional point masses added to the pelvis and femur segments in the EXO condition to account for exoskeleton mass (see Supplemental Material).

Joint kinematics were obtained via inverse kinematics from marker trajectories, and joint moments were computed via inverse dynamics from the joint kinematics and ground reaction forces. Signals were low-pass filtered with zero-lag, fourth-order Butterworth filters (markers: 15~Hz; ground reaction forces: 20~Hz; resulting kinematics/moments: 6~Hz). Data were segmented into strides using HS events detected from the vertical ground reaction force and time-normalized by resampling each stride to a fixed number of samples.

Ground reaction force and center-of-pressure signals were inspected for integrity. For treadmill trials (LG, RA, RD), the left tread data were excluded due to an offset in force/center-of-pressure measurements; therefore, analyses used the participant’s right side for LG and RA, and their left side for RD (walking in the opposite direction). For overground SA/SD and STS, both sides were analyzed. Note that participants exhibited no observable left/right bias, which was confirmed by comparing bilateral kinematics (see Supplemental Material).

The biological hip moment was computed by subtracting the time-synchronized exoskeleton torques (logged on the Raspberry Pi 5) from the net hip moment obtained from inverse dynamics in the EXO condition. In the No EXO condition, the inverse-dynamics hip moment equals the biological hip moment. For the knee and ankle, inverse-dynamics moments equal biological joint moments in both conditions because no external torques were applied at those joints. Moment and power profiles were averaged across all gait cycles. Peak biological hip moment was defined as the maximum magnitude of the average moment profile, whereas peak biological and total hip power were defined as the maximum positive values of their respective average power profiles.

The positive joint work for each task cycle was obtained by integrating positive joint power, $\max(P(t),0)$, over the task cycle time period. Averaging across all gait cycles gives the average positive work for each task. Summing the average positive biological joint work of hip, knee, and ankle gives the total positive biological lower-limb work.

\subsubsection{Statistical Analysis}

For each participant and task, we computed hip positive work and total lower-limb positive work under two within-subject conditions (No EXO and EXO). As a descriptive check, we assessed whether each outcome was approximately normally distributed using Shapiro--Wilk tests \cite{shapiro1965analysis} (Python 3.11, SciPy 1.10.1).

To test the effect of exoskeleton assistance, we fit separate linear mixed models (LMMs) for each task using restricted maximum likelihood (MATLAB 2024b). In each task-specific model, condition (No EXO vs. EXO) was treated as a categorical fixed effect, and participant was included as a random intercept to account for the repeated-measures structure. 

The linear mixed model is represented by
\begin{equation}
  \begin{split}
    \label{eq:lmm}
    \text{Outcome}\ \sim\  \text{Condition} + \left( 1 | \text{Participant} \right)\ .  
  \end{split}
\end{equation}

For fixed effects, we report the LMM $p$-values and apply Holm--Bonferroni correction to control for multiple comparisons across tasks for the condition effect.

\section{Results}

\subsection{Exoskeleton and Biological Hip Moments}

\begin{figure}[htbp]
\centering
\includegraphics[width = 0.9\linewidth]{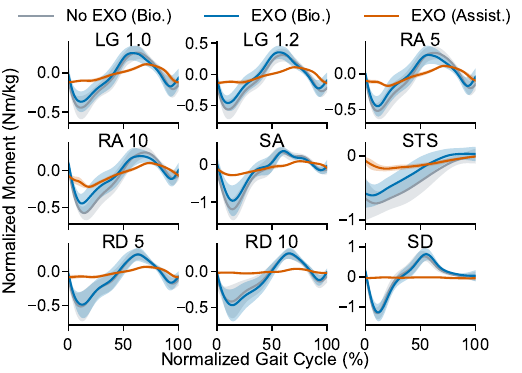}\vspace{-2mm}
\caption{Normalized biological hip moment and exoskeleton assistance torque over a cycle of activities. No EXO (Bio.) stands for the biological hip moment in the No EXO condition, EXO (Bio.) stands for the biological hip moment in the EXO condition, and EXO (Assist.) stands for the commanded torque of the exoskeleton in the EXO condition. Positive values represent flexion moment/torque. Each line represents inter-participant average, and each shaded region represents $\pm 1$ standard deviation.}
\label{fig:torque}
\vspace{-2mm}
\end{figure}

The mass-normalized biological hip moment profiles for the No EXO and EXO conditions, together with the commanded exoskeleton torques, are shown over the activity set in Fig.~\ref{fig:torque}. During hip-intensive activities, the exoskeleton provided substantial assistance in both extension and flexion directions. In contrast, during descent activities (ramp and stair descent), the commanded extension torque was attenuated as desired. This pattern is consistent with the adaptive controller design.

\begin{figure}[htbp]
\centering
\includegraphics[width = 0.9\linewidth]{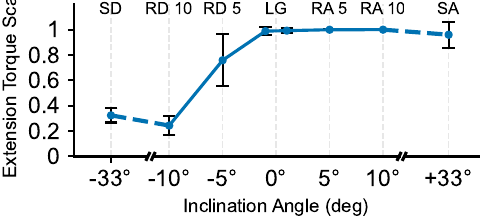}\vspace{-2mm}
\caption{Extension torque scale across activities. Markers represent the inter-participant average scale, and error bars represent $\pm 1$ standard deviation. Note that for SA and SD, the inclination angles are approximated from the horizontal and vertical changes from the bottom to the top of the staircase.}
\vspace{-2mm}
\label{fig:scale}
\end{figure}

To highlight this cross-activity adaptation, Fig.~\ref{fig:scale} shows how the controller’s hip extension torque scales through the attenuation factor $1-\lambda\alpha$ in (\ref{eq:tau_gait}) across incline/decline angles. During descent, extension assistance was reduced, with the lowest scaling observed for stair descent (SD; -33$^\circ$) and steep ramp descent (RD 10; -10$^\circ$). At the shallower decline (RD 5; -5$^\circ$), the extension-torque scale increased but showed greater inter-participant variability. In contrast, near-full extension assistance was provided during level walking at both speeds (LG 1.0 and LG 1.2; 0$^\circ$), ramp ascent at both inclinations (RA 5/RA 10; 5$^\circ$/10$^\circ$), and stair ascent (SA; 33$^\circ$), consistent with the higher hip extension demand of these tasks. Overall, these results reflect the controller objective: to deliver strong assistance during hip-intensive activities while tapering extension torque during descent to promote transparent interaction.

\begin{figure}
    \centering
    \includegraphics[width=0.9\linewidth]{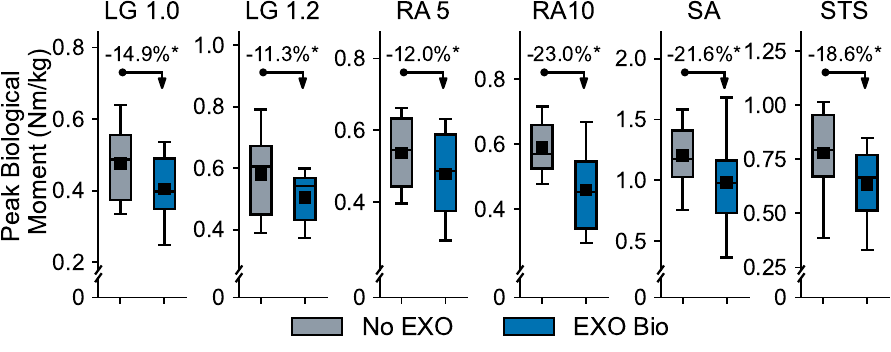}
    \caption{Per-limb, mass-normalized peak biological hip moment across hip-intensive activities under No EXO and EXO conditions. Boxplots show the inter-participant median (horizontal line), interquartile range (box), inter-participant mean (black square), and whiskers to the most extreme observation within 1.5 times of the interquartile range. Numbers indicate the percent reduction in EXO relative to No EXO. Asterisks indicate statistical significance (Holm--Bonferroni-corrected $p \leq 0.05$).}
    \label{fig:peak_moment}
    \vspace{-2mm}
\end{figure}

The mass-normalized peak biological hip moment for hip-intensive activities in both conditions is shown in Fig.~\ref{fig:peak_moment}. In the fitted LMMs, condition had a significant effect on peak hip biological moment for all hip-intensive activities, with hip exoskeleton assistance reducing peak hip biological moment relative to the No EXO condition by 11.3\% to 23.0\%. The greatest reduction occurred during stairs ascent (reduced by 0.23 $\pm$ 0.15 Nm/kg) and the smallest reduction occurred during ramp ascent at 5$^{\circ}$ (reduced by 0.060 $\pm$ 0.04 Nm/kg).

\subsection{Peak Biological and Total Hip Positive Power}

\begin{figure}[htbp]
\centering
\includegraphics[width = 0.9\linewidth]{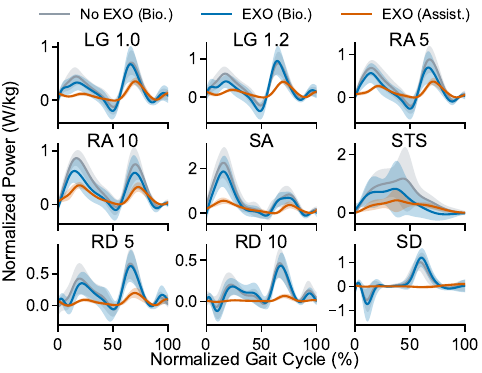}\vspace{-2mm}
\caption{Normalized biological hip power and exoskeleton assistance power over a cycle of activities. No EXO (Bio.) and EXO (Bio.) represent the biological hip power in the No EXO and EXO conditions, respectively, and EXO (Assist.) represents the exoskeleton assistance power in the EXO condition. Lines and shaded regions represent the inter-participant average and $\pm 1$ standard deviation, as in Fig. \ref{fig:torque}.}
\vspace{-2mm}
\label{fig:power}
\end{figure}

Fig.~\ref{fig:power} shows the mass-normalized biological hip power profile for the No EXO and EXO conditions, together with the exoskeleton power profile, for the activity set. Consistent with the torque results, the exoskeleton delivered high assistive power during hip-intensive activities, with prominent positive power peaks in both extension and flexion that aligned with the biological hip positive power peaks \cite{winter2009biomechanics}. In contrast, during descent activities, the extension-related positive power peaks were attenuated, consistent with the reduced exoskeleton extension torque in these tasks. 

Across hip-intensive activities, exoskeleton power closely matched the biological hip power profile, as quantified by the cosine similarity (SIM) in Table~\ref{tab:vivo_SIM} (mean SIM $> 0.8$). During descent activities, SIM decreased as expected, since the controller prioritizes mechanical transparency over maximizing power delivery in these tasks.

\begin{figure}[tbp]
    \centering
    \includegraphics[width=0.95\linewidth]{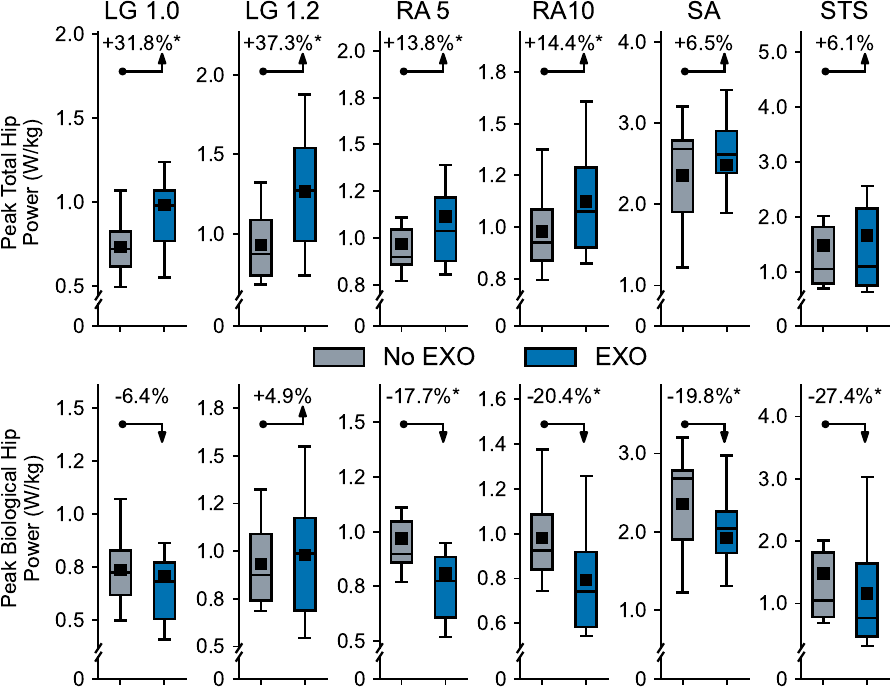}\vspace{-2mm}
    \caption{Per-limb, normalized peak biological hip positive power and total (biological + exoskeleton) hip positive power in No EXO and EXO conditions. Box-and-whisker conventions are as in Fig.~\ref{fig:work}. Numbers denote percent change in EXO relative to No EXO. Asterisks indicate statistical significance (Holm--Bonferroni-corrected $p \leq 0.05$).}
    \vspace{-4mm}
    \label{fig:bio_peak}
\end{figure}

The mass-normalized peak biological hip power and total (biological + exoskeleton) hip positive power for hip-intensive activities in the EXO and No EXO conditions are shown in Fig.~\ref{fig:bio_peak}. For the peak biological hip positive power, in the fitted LMMs, condition had a significant effect during ramp ascent at 5$^\circ$ and 10$^\circ$, stair ascent, and STS, with reductions ranging from 17.7\% to 27.4\%. Condition was not significant for level walking at 1.0 or 1.2~m/s.

In the fitted LMMs for peak total hip positive power, condition had a significant effect during level walking at 1.0 and 1.2~m/s and ramp ascent at 5$^\circ$ and 10$^\circ$, with increases ranging from 13.8\% to 31.8\%. Condition was not significant for stair ascent or STS. 
Individual peak power changes, 95\% confidence intervals (CIs), and Holm--Bonferroni-corrected $p$-values from each LMM are given in Supplemental Material.

\begin{table*}[htbp]
\centering
\caption{Cosine similarity (SIM; average $\pm$ standard deviation) between exoskeleton and biological hip power during the EXO condition. SIM = 1 indicates perfect similarity.}
\label{tab:vivo_SIM}

\setlength{\tabcolsep}{4pt}
\begin{tabular}{c c c c c c c c c c}
\toprule
\textbf{Activity}
& \multicolumn{1}{!{\vrule width 0.4pt}c}{\textbf{LG 1.0}}
& \textbf{LG 1.2}
& \textbf{RA 5}
& \textbf{RA 10}
& \textbf{SA}
& \textbf{STS}
& \textbf{RD 5}
& \textbf{RD 10}
& \textbf{SD} \\
\midrule
\textbf{SIM}
& \multicolumn{1}{!{\vrule width 0.4pt}c}{$0.86 \pm 0.06$}
& $0.83 \pm 0.05$
& $0.88 \pm 0.07$
& $0.94 \pm 0.05$
& $0.91 \pm 0.04$
& $0.91 \pm 0.08$
& $0.80 \pm 0.10$
& $0.73 \pm 0.16$
& $-0.11 \pm 0.43$ \\
\bottomrule
\end{tabular}
\vspace{-4mm}
\end{table*}

\subsection{Biological Joint Positive Work}

\begin{figure}[htbp]
  \centering
  \includegraphics[width = 0.95\linewidth]{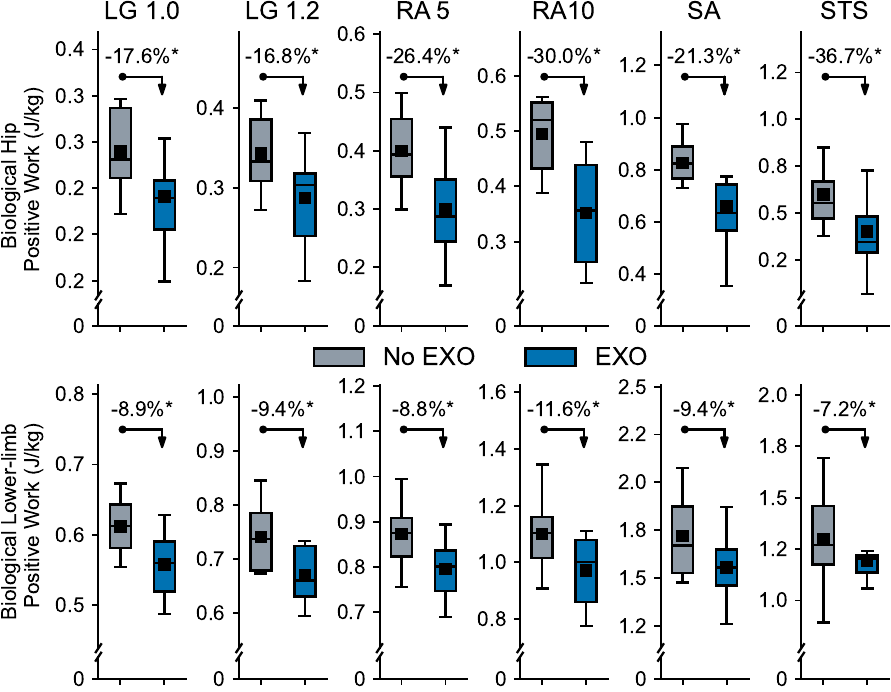}\vspace{-2mm}
  \caption{Per-limb, mass-normalized biological positive work at the hip and for the lower limb (hip + knee + ankle) across hip-intensive activities under No EXO and EXO conditions. Boxplots show the inter-participant median (horizontal line), interquartile range (box), inter-participant mean (black square), and whiskers to the most extreme observation within 1.5 times of the interquartile range. Numbers indicate the percent reduction in EXO relative to No EXO. Asterisks denote significant EXO vs No EXO differences within an activity (Holm--Bonferroni corrected $p \le 0.05$). Task abbreviations (used throughout Figs.~\ref{fig:work}--\ref{fig:power}): LG, level walking; RA5/RA10, ramp ascent at 5$^\circ$/10$^\circ$; SA, stair ascent; STS, sit-to-stand.}
  \vspace{-2mm}
  \label{fig:work}
\end{figure}

The per-limb, mass-normalized biological hip positive work and lower-limb positive work (hip + knee + ankle) for hip-intensive activities (level walking at two speeds, ramp ascent at two inclinations, stair ascent, and STS) in the EXO and No EXO conditions are shown in Fig.~\ref{fig:work}. In the fitted LMMs, condition had a significant effect on hip positive work for all hip-intensive activities, with hip exoskeleton assistance reducing hip positive work relative to the No EXO condition by 16.8\% to 36.7\%. The greatest reduction occurred during STS (0.196 $\pm$ 0.087 J/kg) and the smallest reduction occurred during level walking at 1.0 m/s (0.053 $\pm$ 0.038 J/kg). 

Similarly, condition had a significant effect on lower-limb positive work for all hip-intensive activities, with the exoskeleton condition being 7.2\% to 11.6\% lower than the No EXO condition. The greatest reduction occurred during stair ascent (0.165 $\pm$ 0.103 J/kg) and the smallest reduction occurred during level walking at 1.0 m/s (0.056 $\pm$ 0.034 J/kg). 
% Sex did not have a significant main effect in these models.
Individual results, estimated work changes, 95\% confidence intervals (CIs), and Holm--Bonferroni-corrected $p$-values from each LMM are given in Supplemental Material. No systematic adverse trends or consistent non-responders were observed.

\section{Discussion}

Older adults suffer from hip energetic deficits due to aging, which are exacerbated by the age-related distal-to-proximal shift in joint work demand \cite{devita2000age}. Our objective was to provide multi-task energetic benefits using a versatile hip exoskeleton that delivers task-adaptive assistance during hip-intensive activities while promoting transparent, predictable interactions during descent. The resulting task-agnostic controller matched biological hip power across hip-intensive tasks with an average SIM of 0.8879 (where 1 is perfect similarity). Using this exoskeleton, we observed significant average reductions of 24.7\% in biological hip positive work and 9.3\% in combined lower-limb positive work; a 16.9\% average reduction in peak biological hip moment; a 14.5\% average reduction in peak biological hip positive power; and an 18.3\% average increase in peak total (biological + exoskeleton) hip positive power.

\subsection{Velocity-Modulated Springs for Task-Agnostic Hip Power}
By using velocity-modulated virtual springs, our controller is biomechanically intuitive and inherently sensitive to joint power. This allowed the exoskeleton to meet the energetic objectives under the variable kinematics of older adults. Compared with young adults, older adults exhibit a shift in hip kinematics toward flexion and more sustained hip extension moment and positive power during stance, corresponding to an approximately 8\% delay in zero-crossing timing \cite{devita2000age}. In our results, the biological and exoskeleton torque (Fig.~\ref{fig:torque}) and power (Fig.~\ref{fig:power}) profiles showed more consistent zero-crossing timing, with an approximately 3\% phase delay. Thus, although the controller was optimized and tuned using data from younger populations, its assistance remained appropriately timed and aligned with the movements of the older adults. Overall, the controller achieved high SIM between biological and exoskeleton torque and power, both in silico (Table~\ref{tab:silico_SIM}) and in vivo (Table~\ref{tab:vivo_SIM}), across the hip-intensive tasks. 

The peaks of the exoskeleton torque/power profiles were delayed 50--100~ms compared to biological profiles across the in vivo tasks.
Such delays help synchronize the peaks of assistive torque and joint velocity \cite{dingEffectTimingHip2016}, maximizing mechanical power delivery from the exoskeleton. Accordingly, \cite{molinaro2024} added a fixed delay of 150~ms to the commanded torque based on pilot subject feedback. We instead tuned the slope and offset of the sigmoid function that affects the velocity-based modulation of torque, leading to more task-generalizable phase delays that potentially optimize joint energetic benefits.

Compared with hip-intensive activities, descent tasks (ramp and stair descent) were associated with lower SIM, both in silico and in vivo. This result was expected because the controller's objective during descent was not to maximize assistance but to ensure more transparent interactions. Unlike hip-intensive tasks, descent tasks demand relatively little hip positive power and thus less assistance. Since descent tasks are associated with elevated fall risk in older adults \cite{nevitt1991risk}, We prioritized predictable, low-torque interactions by incorporating an explicit mechanism to attenuate excessive extension torque. This advances our preliminary work \cite{zhang2025task}, which autonomously adjusted controller parameters based on general kinematic cues identified in silico, but produced less consistent assistance in vivo due to the highly variable kinematics of stair descent. Here, we leverage the kinematics at heel-strike (HS), i.e., the inter-thigh angle difference, to naturally distinguish descent from non-descent tasks \cite{posh2025task}. Based on this task context, our approach attenuates only hip extension torque magnitude while preserving the less risk-inducing flexion torque to maintain a uniform controller policy across activities. Stair descent exhibited the lowest SIM because of minor stance-phase torque-shape mismatches, especially in vivo (Table \ref{tab:vivo_SIM}) where measurements did not always match the able-bodied normative dataset used in silico (Table \ref{tab:silico_SIM}). Importantly, stance-phase torque remained negligible (mean torque: -0.023 Nm/kg) in stair descent, providing mechanical transparency. 
Although not tested here, we believe that this transparent behavior has safety and comfort implications in a task category largely overlooked by most hip exoskeleton controllers.

By leveraging the inter-thigh angle difference at HS, we also extended descent modulation (limited to stairs descent in \cite{zhang2025task}) to steep ramp descent, allowing the controller to adapt robustly across all tasks (Fig.~\ref{fig:scale}) while maintaining apt biomimetic assistive power (Fig.~\ref{fig:power}). As shown in Fig.~\ref{fig:scale}, the resulting extension-torque scale decreased substantially during stair and ramp descent, while remaining near unity for level walking and ascent tasks. Notably, the -5$^\circ$ ramp descent condition exhibited larger inter-participant variability, likely because mild descent lies near the transition between level and descent gait strategies \cite{redfern1997biomechanics}: some participants adopted cautious, shortened steps with smaller inter-thigh angle differences at HS (triggering stronger attenuation), whereas others used more level-like kinematics with larger HS inter-thigh angle differences (yielding less attenuation). Consistent with this interpretation, prior work suggests that strategy transitions during ramp descent occur between 9--12$^\circ$ \cite{kawamura1991gait}, which may explain why modulation was more variable at 5$^\circ$ decline than at steeper grades.

In addition to the gait tasks, our controller also adapts its assistance for the important non-gait task of STS, which had the second-highest SIM (average 0.9135) among all tasks. STS is a demanding activity that older adults perform more than 40 times per day on average \cite{dall2010frequency}, yet it is rarely assessed in hip exoskeleton studies. Sit-to-stand involves diverse strategies and kinematics in older adults \cite{schenkman1990whole}, making it difficult to define a single phase variable that generalizes across participants when only torso and thigh kinematics are available, in contrast to gait tasks where phase is more consistently defined \cite{kimWearableHipassistRobot2018}. Its non-cyclic structure also limits adaptive oscillator-based approaches, which typically require multiple steady cycles to converge before delivering consistent assistance \cite{leeGaitPerformanceFoot2017}. In contrast, our controller implements a thigh-velocity-modulated virtual spring that does not rely on gait phase or periodicity, enabling immediate and robust assistance during STS.

\subsection{Exoskeleton Assistance Reduces Joint Mechanical Work}

Our primary hypothesis was supported across all hip-intensive activities (level walking at two speeds, ramp ascent at two inclinations, stair ascent, and STS). In these tasks, assistance significantly reduced biological hip positive work by 16.8--36.7\%, a meaningful effect given the hip's major role in positive work during fast walking and ramp ascent \cite{farris2012mechanics}. This effect is especially relevant for older adults, who have reduced muscular efficiency \cite{hopker2013influence} and exhibit an age-related distal-to-proximal redistribution of lower-limb effort that increases reliance on hip musculature \cite{cofre2011aging}. In \cite{cofre2011aging}, older participants generated 0.036 J/kg and 0.042 J/kg more hip extensor positive work than young participants when walking at 1.0 and 1.3 m/s, respectively. In comparison, our exoskeleton assistance reduced hip positive work by 0.053 and 0.056 J/kg at similar walking speeds (1.0 and 1.2 m/s). This supports our device's potential to increase endurance by reducing muscular effort and delaying the onset of fatigue \cite{green1997mechanisms}, with downstream implications for preventing injuries or falls \cite{worrell1992hamstring, morrison2016walking}. Similarly, during incline walking, older adults compensate for reduced ankle power by increasing their relative hip work by 9.5\% compared to young adults \cite{kuhman2018relationships} (while negative hip work remains relatively small). Given that total lower-limb positive work is comparable between age groups (0.43 J/kg at 5.7$^{\circ}$ and 1.1 m/s \cite{waanders2018advanced}), this shift equates to an estimated $\sim$0.041 J/kg increase in positive hip work in older adults. Crucially, the positive work provided by our exoskeleton (0.096 J/kg at 5$^\circ$ and 1.0 m/s) more than compensates for this extra burden.

Beyond the hip, assistance reduced combined lower-limb positive work across all hip-intensive activities by 7.2--11.6\%. This overall reduction was driven primarily by the large decrease in biological hip positive work, dominating smaller and more variable changes in knee and ankle work across tasks (see Supplemental Material). The weight of the exoskeleton appeared to increase demand on the adjacent joints during tasks like STS, which exhibited a higher sagittal moment at the knee. Nevertheless, statistical analyses found no significant EXO-related changes in knee or ankle positive work across tasks (see Supplemental Material), indicating that the lower-limb work reduction was not primarily due to redistribution to the distal joints.
This pattern matches prior findings in young adults, where a neural network-based task-agnostic hip exoskeleton reduced hip work with mixed knee and ankle responses during level and incline walking \cite{molinaro2024}. Our lower reductions in positive work metrics compared to the young healthy cohort in \cite{molinaro2024} may reflect population differences and/or the longer adaptation times that older adults need to fully benefit from assistance \cite{lakmazaheri2024optimizing}. The hip-knee exoskeleton in \cite{molinaro2024a} had no significant effect on lower-limb positive work during stair ascent, whereas our hip exoskeleton caused the greatest reduction during this task. A plausible explanation is the added distal mass from the knee module in \cite{molinaro2024a}, which can impose a larger weight (and inertia) penalty during dynamic tasks such as stair ascent \cite{mooneyAutonomousExoskeletonReduces2014}. Given the established role of joint-level mechanical work in the walking metabolic cost \cite{silder2012predicting, donelan2002mechanical}, our holistic lower-limb positive work reductions may indicate metabolic and endurance benefits, though these assessments were outside the scope of this joint-level study. 

By offloading positive hip work across hip-intensive tasks, our task-agnostic exoskeleton directly mitigates the increased demand older adults place on their hips \cite{delabastita2021distal, devita2000age}, while simultaneously reducing overall lower-limb positive work. This is important to ensuring the hip benefits do not come at a significant cost to the weakened distal joints and overall effort during community ambulation.

\subsection{Exoskeleton Assistance Supports Peak Joint Power}

Exoskeleton assistance significantly increased total (exoskeleton + biological) peak hip positive power during level walking (two speeds) and ramp ascent (two inclinations), tasks in which the hip is a primary contributor to propulsion \cite{farris2012mechanics}. For the remaining hip-intensive tasks, the changes were directionally consistent but did not reach statistical significance. Because older adults exhibit substantial age-related reductions in hip power \cite{dean2004age} that are closely linked to functional performance (e.g., walking speed) \cite{graf2005effect}, increasing total hip power may translate to improved mobility during hip-demanding activities. Notably, during fast walking the exoskeleton increased total peak hip positive power by 0.334~W/kg. For context, 10 weeks of leg power training improved peak hip extension power during fast walking by 0.2~W/kg \cite{beijersbergen2017hip}. Our exoskeleton thus produced a larger acute benefit under comparable conditions, suggesting a potentially efficient intervention for hip power deficits. That said, the delayed exoskeleton power shown in Fig.~\ref{fig:power}, while potentially beneficial for reducing mechanical work, can decrease peak total power and contribute to the non-significant effects observed during stair ascent and STS. Furthermore, these tasks also show greater inter-participant variability due to differing movement strategies in older adults \cite{reevesOlderAdultsEmploy2009, schenkman1990whole}, which may further reduce statistical power. Nevertheless, these findings suggest that our hip exoskeleton can increase total hip power capacity in older adults, particularly during hip-dominant locomotor activities.

Exoskeleton assistance significantly reduced peak biological hip power during ramp ascent, stair ascent, and STS, likely reflecting the corresponding reductions in peak biological hip moment. (Power changes during the two level walking tasks were not statistically significant.) In older adults, the distal-to-proximal shift increases reliance on hip power to compensate for reduced ankle power generation \cite{franz2014advanced}. Coupled with age-related declines in hip power capacity, this can create a demand-capacity mismatch that forces older adults to perform ADLs closer to their maximal capacity, particularly during stair negotiation and rising from a chair \cite{hortobagyi2003old}. Here, assistance reduced peak biological hip power by 19.8\% during stair ascent and 27.4\% during STS, potentially preserving hip power reserve for balance control, where the hip plays an important role \cite{jeon2019muscle}. By lowering hip power demand (i.e., effectively increasing functional power capacity), the exoskeleton may help lower the ``disability threshold'' for hip-intensive activities \cite{fried1997disability}, thus improving the mobility of the elderly population.

\subsection{Limitations and Future Work}
Our primary focus here was on establishing the joint-level energetic effects of our task-agnostic exoskeleton on older adults, and as such we did not evaluate body-level energetic benefits like metabolic cost and functional benefits like walking speed. 
This focus was intentional, as joint energetics is strongly associated with functional status in older adults \cite{foldvari2000association} and directly constrains performance in high-demand transitional activities such as stair ascent and STS, which are critical for mobility independence~\cite{wilken2011role, bassey1992leg}.
Furthermore, while our analyses included numerous activities, they were still limited to a controlled lab setting and may not represent unstructured community environments. Nevertheless, the current results firmly establish our hip exoskeleton prototype as a promising translational strategy to target the specific mechanisms that cause the common mobility deficits in aging populations. Building on this, future work will focus on validating these benefits in community environments and assessing the long-term retention of these energetic benefits. While we kept the controller parameters consistent across participants in this study, leveraging the parametric structure of our task-agnostic controller for human-in-the-loop optimization can further increase the benefits\cite{sladePersonalizingExoskeletonAssistance2022}.

\section{Conclusion}
In this paper, we developed a task-agnostic controller for a hip exoskeleton that targets joint-level energetic benefits in older adults during hip-intensive tasks, while promoting transparent, predictable interactions during descent tasks. We validated the approach in eight older participants across a battery of hip-intensive ADLs. Exoskeleton assistance significantly reduced biological hip positive work and combined lower-limb positive work across all hip-intensive tasks, while providing joint moment and power benefits in most tasks. Collectively, these findings demonstrate that the presented task-agnostic hip exoskeleton can deliver joint-level energetic benefits in older adults, supporting its potential as a biomechanical intervention to offset age-related energetic deficits and improve community mobility.

\section*{Acknowledgment}
The authors thank Emily G. Keller, Victoria Landrum, and Nundini Rawal for assisting experiments and Dr. Corey Powell for statistics consultations. Robert Gregg has an equity interest in Motaris, Inc., which has optioned IP related to this study.

\bibliographystyle{IEEEtran}
\bibliography{bibtex/bib/IEEEexample}

\def\supplementfilename{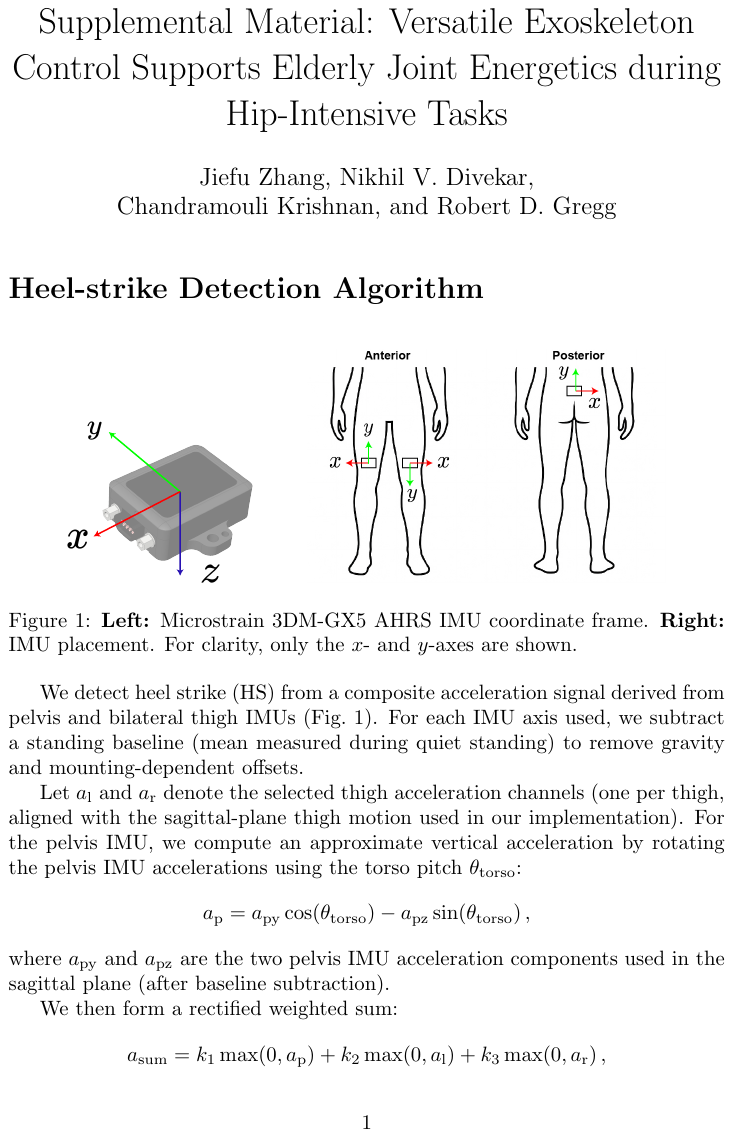}
\pdfximage{\supplementfilename}
\def\numbersupplementpages{\the\pdflastximagepages}

% Loop through and append all pages
\foreach \x in {1,...,\numbersupplementpages} {
    \clearpage
    \includepdf[pages={\x}]{\supplementfilename}
}
% % use section* for acknowledgment
% \section*{Acknowledgment}

\end{document}